\documentclass[letterpaper, 10 pt, conference]{ieeeconf}  

\IEEEoverridecommandlockouts                              
                                                          
\UseRawInputEncoding
\usepackage{url}
\usepackage{comment}
\usepackage{amsmath,amssymb,amsfonts, mathtools}
\usepackage{graphicx}
\usepackage{bm}
\usepackage{epstopdf}
\usepackage{placeins}
\usepackage{bm}
\usepackage{cite}
\let\labelindent\relax
\usepackage[inline]{enumitem}
\usepackage{multicol}
\usepackage{tikz}
\usepackage{caption}
\usepackage{subcaption}
\usepackage{soul}
\usepackage{wrapfig}
\usepackage[normalem]{ulem}
\useunder{\uline}{\ul}{}
\usepackage[inline]{enumitem}
\usepackage{booktabs}

\newcommand{\vect}[1]{\boldsymbol{#1}}

\title{\LARGE \bf Learning from Successful and Failed Demonstrations via Optimization}

\author{Brendan Hertel and S. Reza Ahmadzadeh
	\thanks{Persistent Autonomy and Robot Learning (PeARL) Lab, University of Massachusetts Lowell, Lowell, MA, 01854. Email: {\tt\small brendan\_hertel@student.uml.edu,reza@cs.uml.edu}}}

\begin{document}

\maketitle
\thispagestyle{empty}
\pagestyle{empty}

\begin{abstract}
Learning from Demonstration (LfD) is a popular approach that allows humans to teach robots new skills by showing the correct way(s) of performing the desired skill. Human-provided demonstrations, however, are not always optimal and the teacher usually addresses this issue by discarding or replacing sub-optimal (noisy or faulty) demonstrations. We propose a novel LfD representation that learns from both successful and failed demonstrations of a skill. Our approach encodes the two subsets of captured demonstrations (labeled by the teacher) into a statistical skill model, constructs a set of quadratic costs, and finds an optimal reproduction of the skill under novel problem conditions (i.e. constraints). The optimal reproduction balances convergence towards successful examples and divergence from failed examples. We evaluate our approach through several 2D and 3D experiments in real-world using a UR5e manipulator arm and also show that it can reproduce a skill from only failed demonstrations. The benefits of exploiting both failed and successful demonstrations are shown through comparison with two existing LfD approaches. We also compare our approach against an existing skill refinement method and show its capabilities in a multi-coordinate setting.
\end{abstract}

\section{Introduction}
\label{sec:intro}

As robots become more popular and enter the hands of less experienced users, the ability for users to incorporate desired skills (including primitive actions) into a robot must become more reliable and robust. A natural method for teaching new skills to robots without programming is Learning from Demonstration (LfD), in which a human teacher provides the robot with multiple demonstrations of a skill~\cite{Argall2009survey}. Given the set of examples, the robot builds a skill model that can potentially reproduce the skill even in novel situations. However, most existing LfD approaches rely on optimal or near-optimal human-provided demonstrations. It has been shown that the efficiency and robustness of state-of-the-art LfD algorithms change drastically when employed by users with various levels of robotics experience~\cite{rana2020benchmark}.

Conventionally, this issue has been addressed by saving near-optimal demonstrations and discarding or replacing sub-optimal ones through comparison and evaluation. The effectiveness of this process strongly depends on the user's level of expertise and their prior knowledge of the LfD algorithms. Several LfD approaches try to handle the issue of sub-optimal demonstrations through skill refinement either by assigning an importance value to each demonstration~\cite{argall2010tactile}, or by calculating and imposing geometric constraints to the demonstration space~\cite{Ahmadzadeh2018TLGC}. Other approaches use reinforcement learning (RL) to improve a skill model constructed from sub-optimal demonstrations~\cite{kormushev2010pancake}. Approaches that combine LfD and RL usually require an excessive number of iterations and need an effective, well-defined reward function. 

Furthermore, almost all these approaches have been designed to learn skill models from successful examples that provide information about \emph{what-to-imitate}. However, they ignore failed examples of the skills that provide information about \emph{what-not-to-imitate}. There are few existing approaches that learn from failed demonstrations. Of these approaches, some rely solely on failed demonstrations~\cite{grollman2011donut} while others rely on extra knowledge (in the form of features) provided by the user~\cite{shiarlis2016inverse}.

\begin{figure}[t]
\centering
\includegraphics[trim=0 0em 0 0, clip, width=0.50\columnwidth]{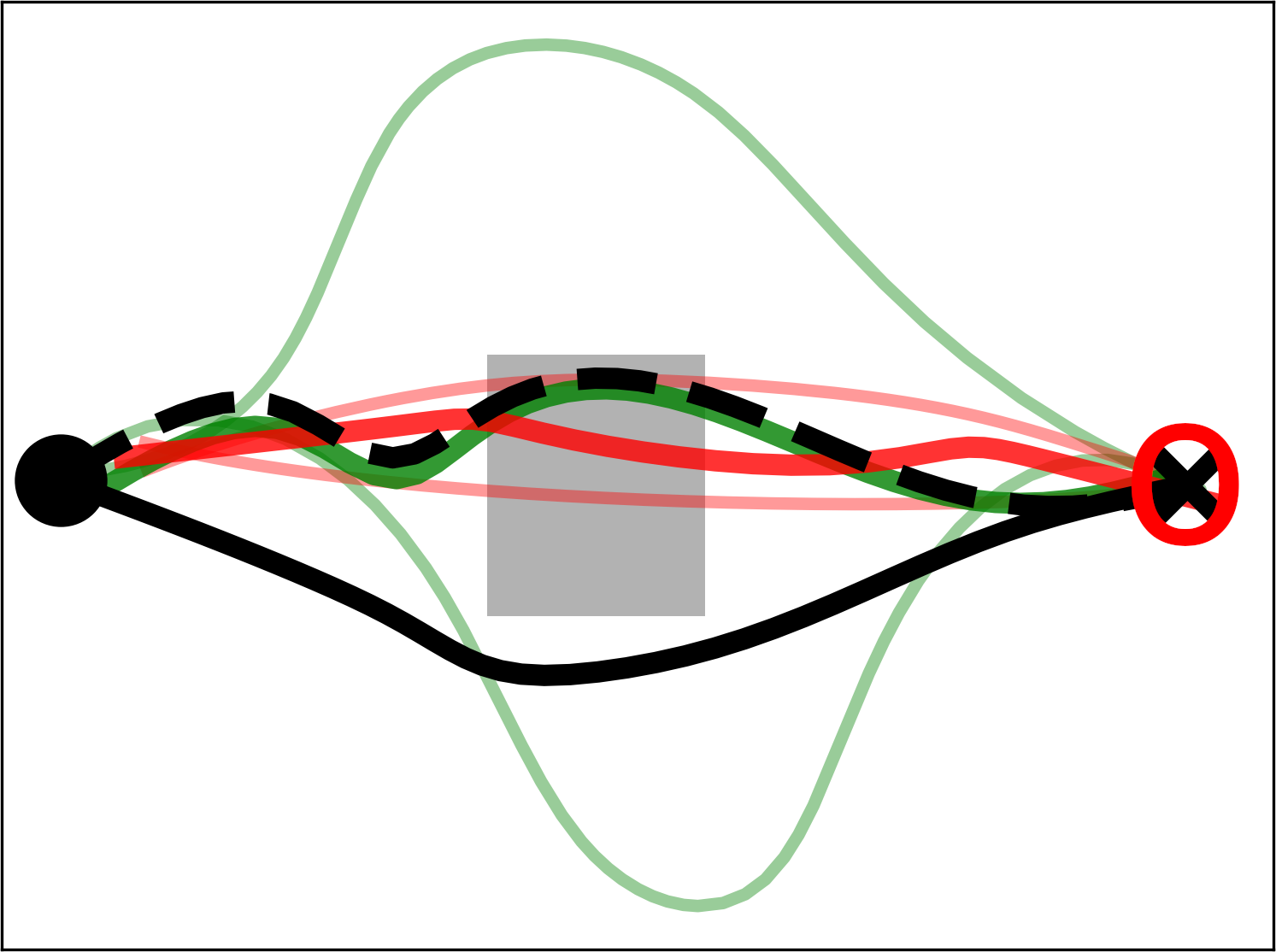}
\includegraphics[trim=0 0em 0 0, clip, width=0.473\columnwidth]{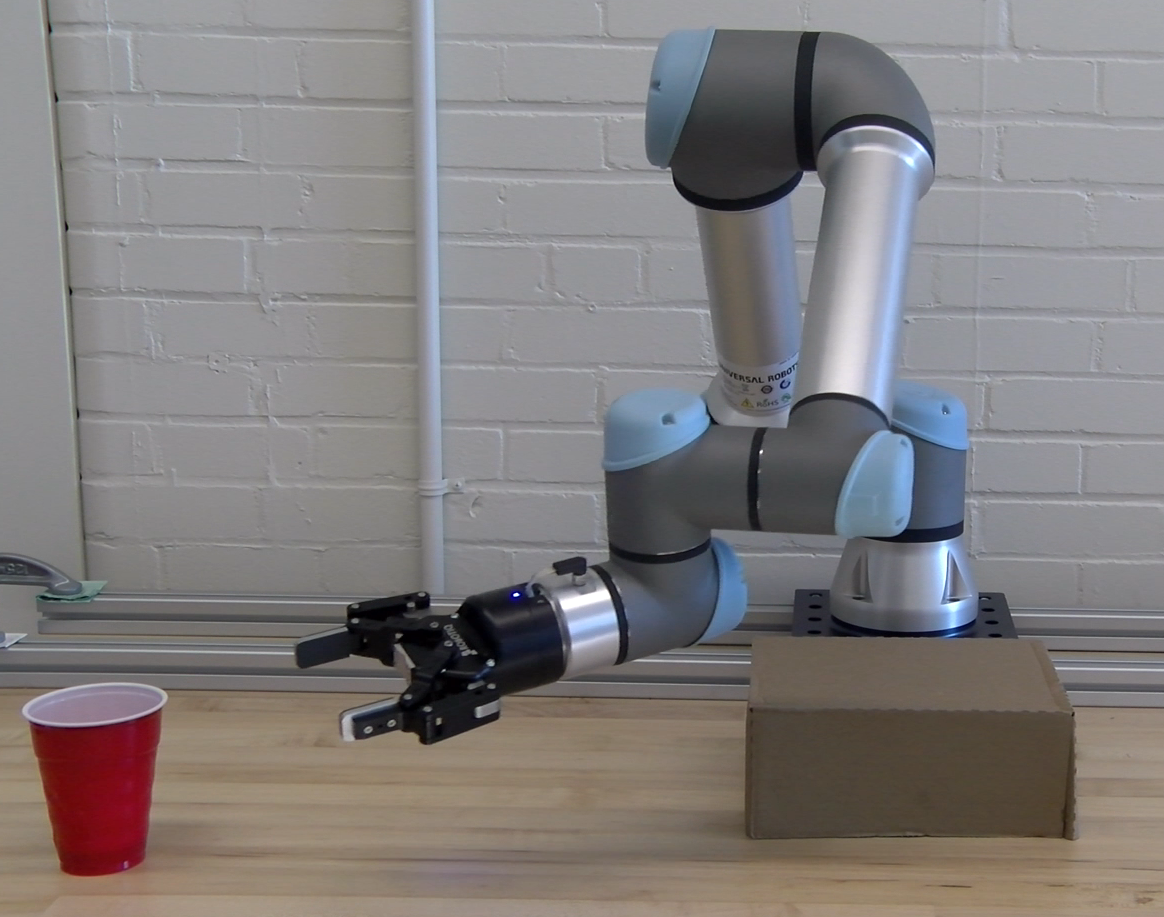}
\caption{\small{Learning from successful (green) and failed (red) demonstrations results in reproductions that can avoid obstacles by diverging from the failed demonstration sets (details in Sec.~\ref{sec:Exp_B}).}} \label{fig1}
\end{figure}

In this paper, we propose a novel LfD approach, \textit{Trajectory Learning from Failed and Successful Demonstrations (TLFSD)}, that can learn from both successful (near optimal) and failed (sub-optimal) demonstrations of a skill. Given the two sets of labeled demonstrations, our approach estimates the joint distributions of each set and forms quadratic costs which represent deviation from successful demonstrations and convergence to failed demonstrations. Given a set of novel skill constraints (e.g., initial, final, and via points), our approach finds an optimal reproduction of the skill satisfying those conditions. An application of the proposed method is shown in Fig.~\ref{fig1}. Another advantage of our approach is that it can reproduce a skill in cases where either the successful or failed demonstration set is empty. Additionally, the algorithm can perform iterative skill refinement by adding the labeled reproductions to the demonstration sets. We evaluate our approach through several 2D and 3D real-world experiments using a UR5e manipulator arm. We also compare our approach to three LfD approaches, one of which was designed for skill refinement.

\section{Related Work}
\label{sec:RW}

To generate successful reproductions of a skill, many LfD approaches rely on optimal or near-optimal demonstrations. Since the demonstrations provided by inexperienced users are not always optimal, the obtained models also generate sub-optimal reproductions~\cite{rana2020benchmark}. To address this issue, some approaches use skill refinement techniques which allow for small corrections in demonstrations and/or the learned skill models~\cite{argall2010tactile, rana2018learning, Ahmadzadeh2018TLGC}. Argall et al. proposed an iterative approach that increases the quality of reproductions by assigning higher weights to newer \emph{more optimal} demonstrations and forgetting older \emph{less optimal} ones~\cite{argall2010tactile}. Alternatively, Ahmadzadeh et al. proposed a skill refinement method that allows the user to kinesthetically refine a sub-optimal demonstration or reproduction and remodel the skill by imposing a set of geometric constraints on the skill model~\cite{Ahmadzadeh2018TLGC}. The probabilistic approach proposed by Rana et al. constructs a skill model that weights demonstrations according to a posterior distribution (e.g., representing distance to an obstacle) and can generate reproductions which are more likely to achieve success~\cite{rana2018learning}. Unlike these methods, our approach can learn from both successful and failed examples of the desired skill and does not require a weighting process.  

Another family of approaches benefits from combining LfD with Reinforcement Learning ~\cite{kormushev2010pancake}. Given a set of sub-optimal demonstrations, these approaches first construct a skill model using an LfD representation. A reinforcement learning (RL) algorithm then uses human guidance in the form of a reward function to improve the skill model and reproductions of the skill gradually. One of the drawbacks of these approaches is that the definition of an effective reward function is a nontrivial task, especially by inexperienced users. Another shortcoming is that RL algorithms require many iterations to find a solution in continuous state-action spaces which is time-consuming and usually impractical in the real-world. Our proposed approach, on the other hand, does not require excessive iterations and only requires labeled demonstrations (as either failed or successful) which is more trivial than defining a reward function. 

One of the few approaches that learn from failed demonstrations was proposed by Grollman and Billard~\cite{grollman2011donut}. Given a set of failed demonstrations, their algorithm builds a statistical model using GMM/GMR~\cite{Calinon2007GMM}, generates new trajectories by exploring the demonstration space, and iteratively converges to a successful solution. They define failed demonstrations as close attempts at completing the desired task, which may not always be the case. Based on this assumption, the algorithm only relies on failed demonstrations and ignores successful demonstrations. Another drawback of the algorithm is that it encodes the skill using position and velocity information and as a result, the reproduced trajectories might include high-velocity values. Our approach allows for the encoding of successful and failed demonstrations, and we show that it can learn a successful reproduction by starting from only a set of failed demonstrations.

Another group of approaches relies on inverse reinforcement learning to compute a reward function from a set of demonstrations. An RL algorithm can then construct or improve a skill model based on the estimated reward function. Shiarlas et al. proposed Inverse Reinforcement Learning from Failure (IRLF) that can discover (i.e., fit) features present in both successful and failed demonstrations, and use them to reproduce the skill~\cite{shiarlis2016inverse}. One of the disadvantages of IRLF is that the set of features must be defined by the user. Additionally, IRLF requires multiple demonstrations, whereas we show that our approach can find a successful solution using a single failed or successful demonstration.

\begin{figure}[t]
\centering
\vspace{0.6em}
\includegraphics[trim=0 0em 0 0, clip, width=0.48\textwidth]{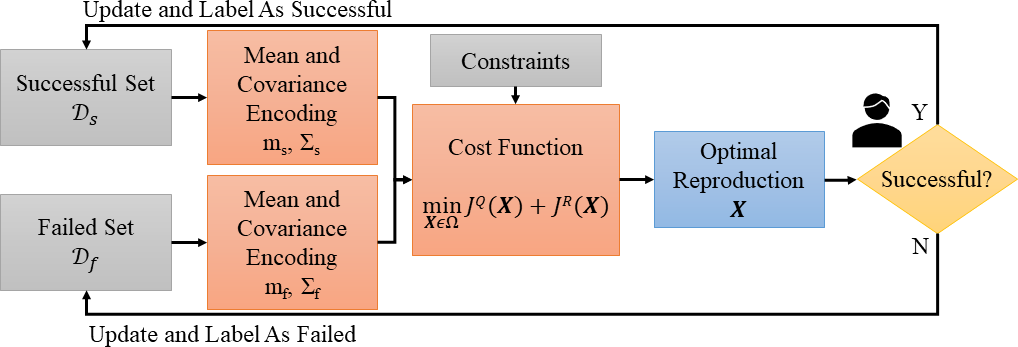}
\caption{\small{Workflow of the proposed approach.}} \label{basic_flow}
\end{figure}

\section{Methodology}
\label{sec:Method}

In this section, we describe the technical details of our approach and its workflow as illustrated in Fig. \ref{basic_flow}.

\subsection{Capturing and Labeling Demonstrations}

The algorithm starts by receiving a set of labeled demonstrations $\mathcal{D}$ that includes two classes of successful and failed demonstrations, $\mathcal{D}=\{\mathcal{D}_s, \mathcal{D}_f \}$. We assume these subsets include $M$ and $N$ demonstrations denoted by $\mathcal{D}_s = \{\vect{X}_{s}^{1}, \dots, \vect{X}_{s}^{M}\}$ and $\mathcal{D}_f = \{\vect{X}_{f}^{1}, \dots, \vect{X}_{f}^{N}\}$, respectively. Later, we show that our approach can handle cases where either one of these subsets is empty. 

Each demonstration in $\mathcal{D}$ is a discrete finite-length trajectory $\vect{X}^j = [\vect{x}_1, \vect{x}_2, ..., \vect{x}_T]^{\top} \in \mathbb{R}^{T \times n}$ in robot task space where $\vect{x}_t = [x_{t(1)}, x_{t(2)}, ..., x_{t(n)}] \in \mathbb{R}^n$ is a single $n$-D observation at time $t$. We use kinesthetic teaching for capturing demonstrations, however, any other demonstration technique such as shadowing and teleoperation can also be used. We then align the raw demonstrations through interpolation and resampling. Other methods such as Dynamic Time Warping (DTW)~\cite{Berndt1994DTW} also could be utilized. We also assume that the human teacher constructs the dataset $\mathcal{D}$ by labeling each collected demonstration either as successful or failed.  


\subsection{Encoding Demonstrations into Costs}
\label{sec:encoding}

After constructing the demonstration set $\mathcal{D}$, we estimate the joint distribution $p(t,x)$ between spatial samples $x$ and time $t$ for each subset $\mathcal{D}_s$ and $\mathcal{D}_f$ independently. Although these joint distributions can be estimated using various probabilistic approaches, we use a mixture model of $K$ $n$-dimensional Gaussian components defined by the joint distribution $p(t,x)=\sum_{k=1}^K \pi^k p(t,x|k)$, where $\pi^k$ and $p(t,x|k) \sim \mathcal{N}(.;\mu^k , \Sigma^k)$ denote the prior and conditional distribution, respectively. We encode each subset of temporally-aligned demonstrations as a Gaussian mixture model (GMM) for which the prior $\pi^k$, mean $\hat{\mu}^k = [\mu_t^k,
\mu_x^k]^{\top}$ and conditional covariances $\hat{\Sigma}^k = 
\begin{bmatrix}
\Sigma_{t,t} & \Sigma_{t,x} \\
\Sigma_{x,t} & \Sigma_{x,x}
\end{bmatrix}$ can be estimated using the Expectation-Maximization algorithm~\cite{Calinon2007GMM}. This process results in two sets of GMM parameters $\mathcal{P}_s = \{ \pi_s^1, \dots, \pi_s^{K_s}; \hat{\mu}_s^1, \dots, \hat{\mu}_s^{K_s}; \hat{\Sigma}_s^1, \dots, \hat{\Sigma}_s^{K_s}\}$ and $\mathcal{P}_f = \{ \pi_f^1, \dots, \pi_f^{K_f}; \hat{\mu}_f^1, \dots, \hat{\mu}_f^{K_f}; \hat{\Sigma}_f^1, \dots, \hat{\Sigma}_f^{K_f}\}$ representing the maximum likelihood solution for the corresponding subset of demonstrations. The optimum number of Gaussian components for each subset ($K_s$ and $K_f$) that maximizes the likelihood while minimizing the number of parameters can be determined using Bayesian Information Criterion (BIC)~\cite{schwarz1978estimating}.

Given a time vector and a learned GMM parameter set, like~\cite{Calinon2007GMM}, we can use Gaussian mixture regression (GMR) to construct a mean value $m_t = \mathbb{E}[x|t]$ and a corresponding covariance matrix $\Sigma_t=\text{Var}[x|t]$ at each time step $t$. The mean and covariances can be calculated as $m_t = \sum_{k=1}^{K} \beta^k (\mu^k + \Sigma^k_{x,t} (\Sigma^k_{t,t})^{-1} (t-\mu_t^k))$ and $\Sigma_m = \sum_{k=1}^{K} (\beta^k)^2 (\Sigma_{x,x} - \Sigma_{x,t} (\Sigma_{t,t})^{-1} \Sigma_{t,x})$, where $\beta^k = p(t|k) / \sum_{l=1}^K p(t|l)$.
We use a single time vector and apply GMR to the learned GMM parameter sets $\mathcal{P}_s$ and $\mathcal{P}_f$ independently. As a result, we obtain two mean trajectories $\vect{m}_s$ and $\vect{m}_f$ together with their corresponding covariance matrices $\vect{\Sigma}_f$ and $\vect{\Sigma}_s$. 

In the next step, we use the estimated joint distributions to construct two quadratic cost functions $J^Q_s$ and $J^Q_f$ using: 
\begin{align}
    J^Q_i(\vect{X}) &= (\vect{X} -{\vect{m}}_{i})^\top {\vect{\Sigma}}_{i}^{-1} (\vect{X} - {\vect{m}}_{i}), i \in \{s,f\}.
\end{align}

\subsection{Cost Scalarization}

To reproduce trajectories that resemble successful demonstrations, our approach penalizes deviations from the conditional mean of the successful subset by minimizing  $J^Q_s$, and rewards deviations from the conditional mean of the failed subset by maximizing $J^Q_f$. The combined cost can be obtained through scalarization as $J^Q = J^Q_s - J^Q_f$.   

\begin{figure}[t]
\centering
\vspace{0.6em}
\includegraphics[trim=0 0em 0 0, clip, width=0.85\columnwidth]{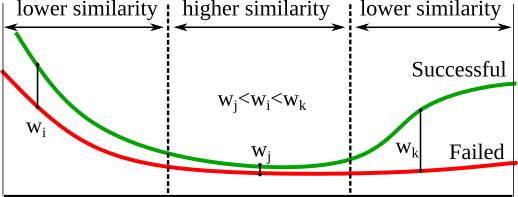}
\caption{\small{Similarities between successful and failed demonstrations vary locally. 
}} \label{curve_sim}
\end{figure}

However, when a human teacher labels a demonstration as failed, it might be because of partial dissimilarities. Fig.~\ref{curve_sim} illustrates two demonstrations of a skill, one labeled as successful and the other as failed. It can be seen that the two demonstrations include segments with higher and lower similarity values. Therefore, when combining the quadratic costs, we consider local importance of dissimilarities represented in $J^Q_f$ by measuring distance between discrete points of the two trajectories. Common similarity metrics such as $L_p$-norms and DTW~\cite{Berndt1994DTW} can be used to measure the distance. If the trajectories consist of different numbers of points, Euclidean distance with a sliding window can be used~\cite{su2020survey}. For dense trajectories, Piece-wise DTW~\cite{keogh2000scaling} can be employed that operates on a higher-level abstraction of data. In this paper, we generate the conditional means, $\vect{m}_s$ and $\vect{m}_f$, with the same number of points and then use the $L_2$-norm to build a dissimilarity vector $\vect{w}_{\text{sim}}$. The combined quadratic cost can then be formed as
\begin{equation}
    J^Q = J^Q_s - \vect{w}_{\text{sim}} J^Q_f, \label{eq:q}
\end{equation}
\noindent where $\vect{w}_{\text{sim}} = \|\vect{m}_{s} - \vect{m}_{f}\|_2$.

In addition to the quadratic terms in $J^Q$, we use a regularizer term, $J^R$, that represents elastic energy~\cite{gorban2005elastic}. Inclusion of the regularizer improves the smoothness of the reproduced trajectory by penalizing high elastic energy which is calculated based on the total length and non-equidistant distribution of points along the trajectory. The elastic energy cost function can be written as 
\begin{align}
J^R(\vect{X}) = \frac{\lambda}{2} \delta(\vect{X})^{\top}\delta(\vect{X}), \label{eq:reg}
\end{align} 
\noindent where $\lambda$ is the spring constant parameter (also known as the regularization factor), and $\delta(\vect{x}_i) = \vect{x}_i - \vect{x}_{i-1}$ denotes the displacement. While large values of $\lambda$ result in a straight line, small values of $\lambda$ produce jagged and noisy trajectories. 

\subsection{Skill Reproduction}

To find an optimal reproduction of the demonstrated skill, we formalize the optimization problem using~\eqref{eq:q} and~\eqref{eq:reg} as 
\begin{align}
    &\underset{\vect{X} \in \Omega}{\text{minimize }} J^Q(\vect{X}) + J^R(\vect{X}), \label{eq:opt}
\end{align}

\noindent where $\Omega = \{ \vect{X} | C_i(\vect{X}) = 0, i \in \mathcal{E}\}$ denotes the feasible set that includes the set of points $\vect{X}$ (solutions) that satisfy equality constraint functions $C_i$, and $\mathcal{E}$ denotes the set of finite indices. The equality constraints $C_i(\vect{X}) = 0$ represent the boundary conditions of the problem and allows us to reproduce trajectories conditioned on unforeseen initial, final, and via points. The constrained optimization problem in~\eqref{eq:opt} can be transformed into an unconstrained optimization problem using methods such as quadratic penalty, logarithmic barrier, and augmented Lagrangian technique~\cite{nocedal2006numerical}. In this paper, we use the quadratic penalty method that adds one quadratic term for each constraint to the original objective. The resulting optimization problem can be written as
\begin{align}
    &\underset{\vect{X} \in \mathbb{R}}{\text{minimize }} J^Q(\vect{X}) + J^R(\vect{X}) + \frac{1}{2 \rho} \sum_{i \in \mathcal{E}} C_i^2(\vect{X}), \label{eq:opt2}
\end{align}

\noindent where $\rho$ is a constant denoting the effect of constraints. Smaller values of $\rho$ penalize deviation from constraints more strongly\footnote{Available at \url{https://github.com/brenhertel/TLFSD}}. 

\section{Experiments}
\label{sec:Experiments}

\begin{figure*}[ht]
\centering
\vspace{0.6em}
\includegraphics[trim=0 0em 0 0, clip, width=0.24\textwidth]{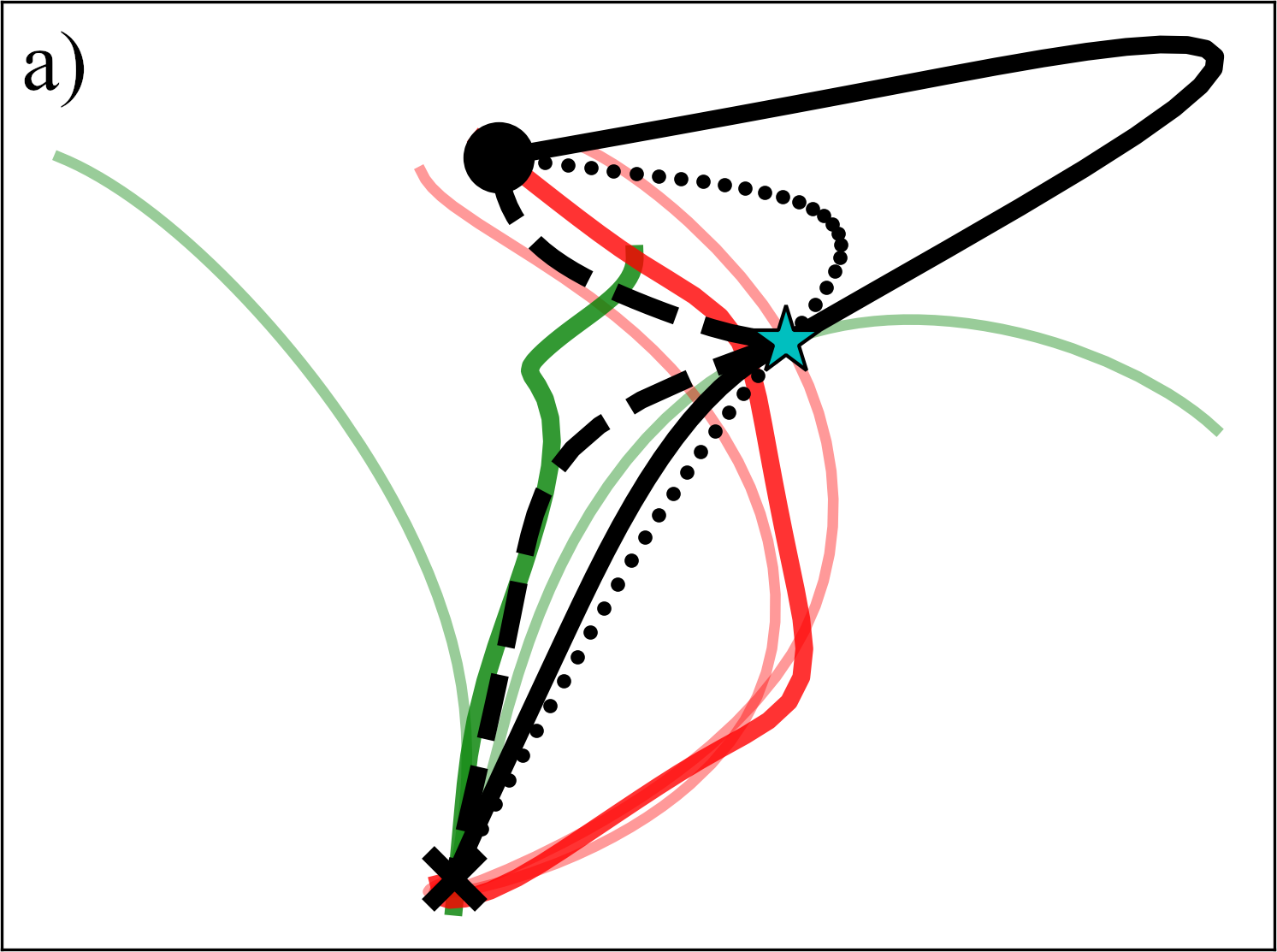}
\includegraphics[trim=0 0em 0 0, clip, width=0.24\textwidth]{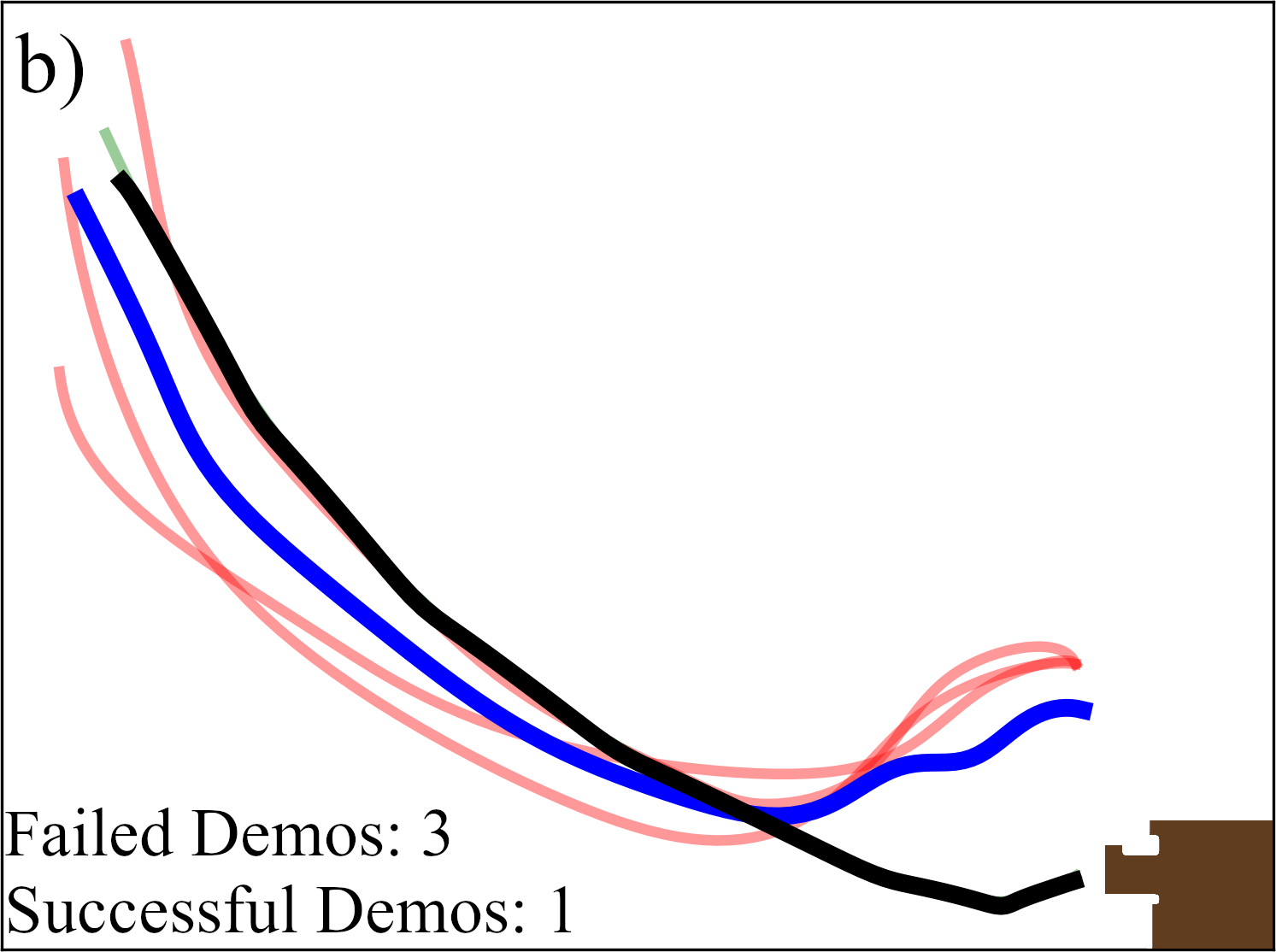}
\includegraphics[trim=0 0em 0 0, clip, width=0.24\textwidth]{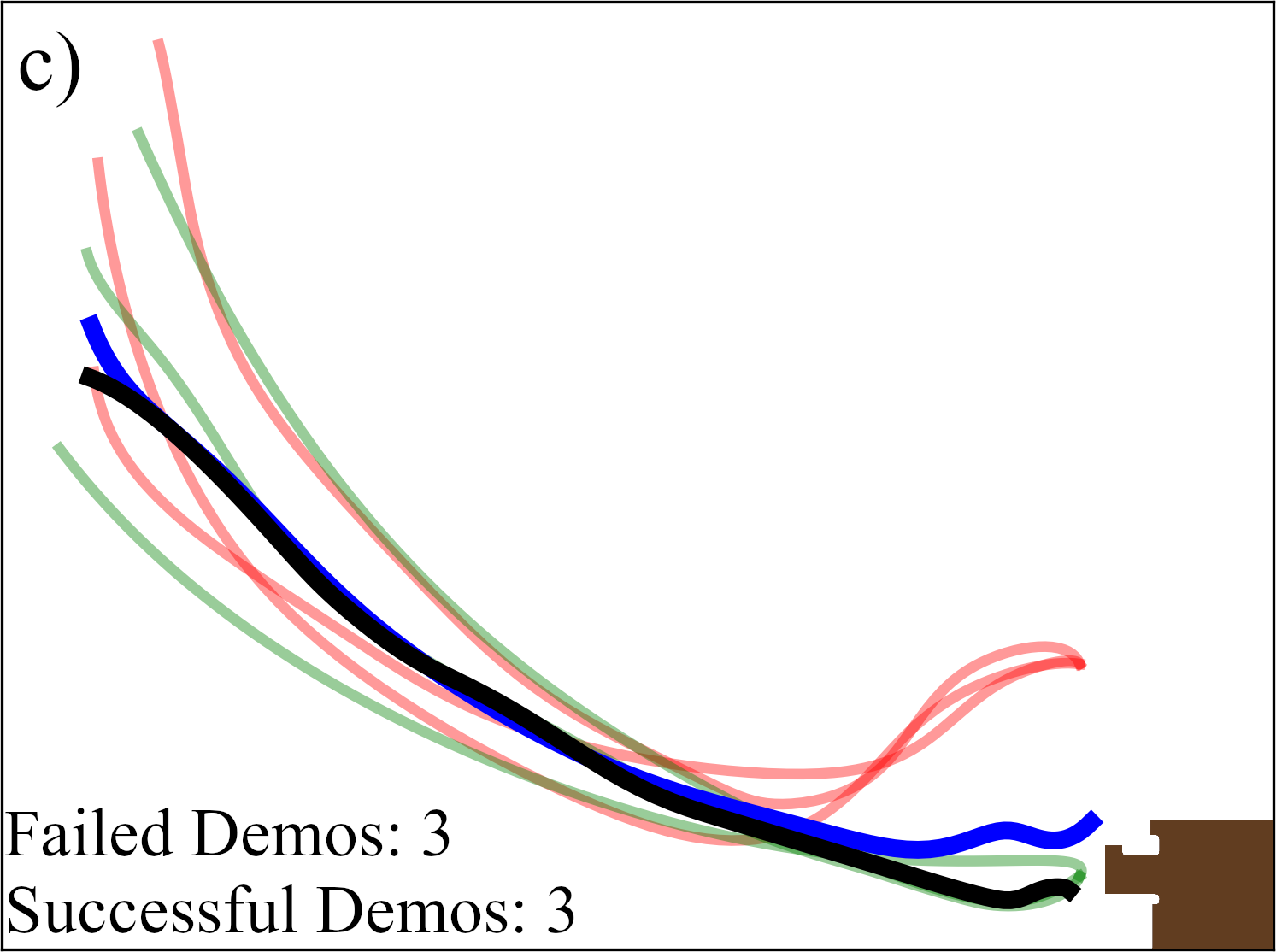}
\includegraphics[trim=0 0em 0 0, clip, width=0.24\textwidth]{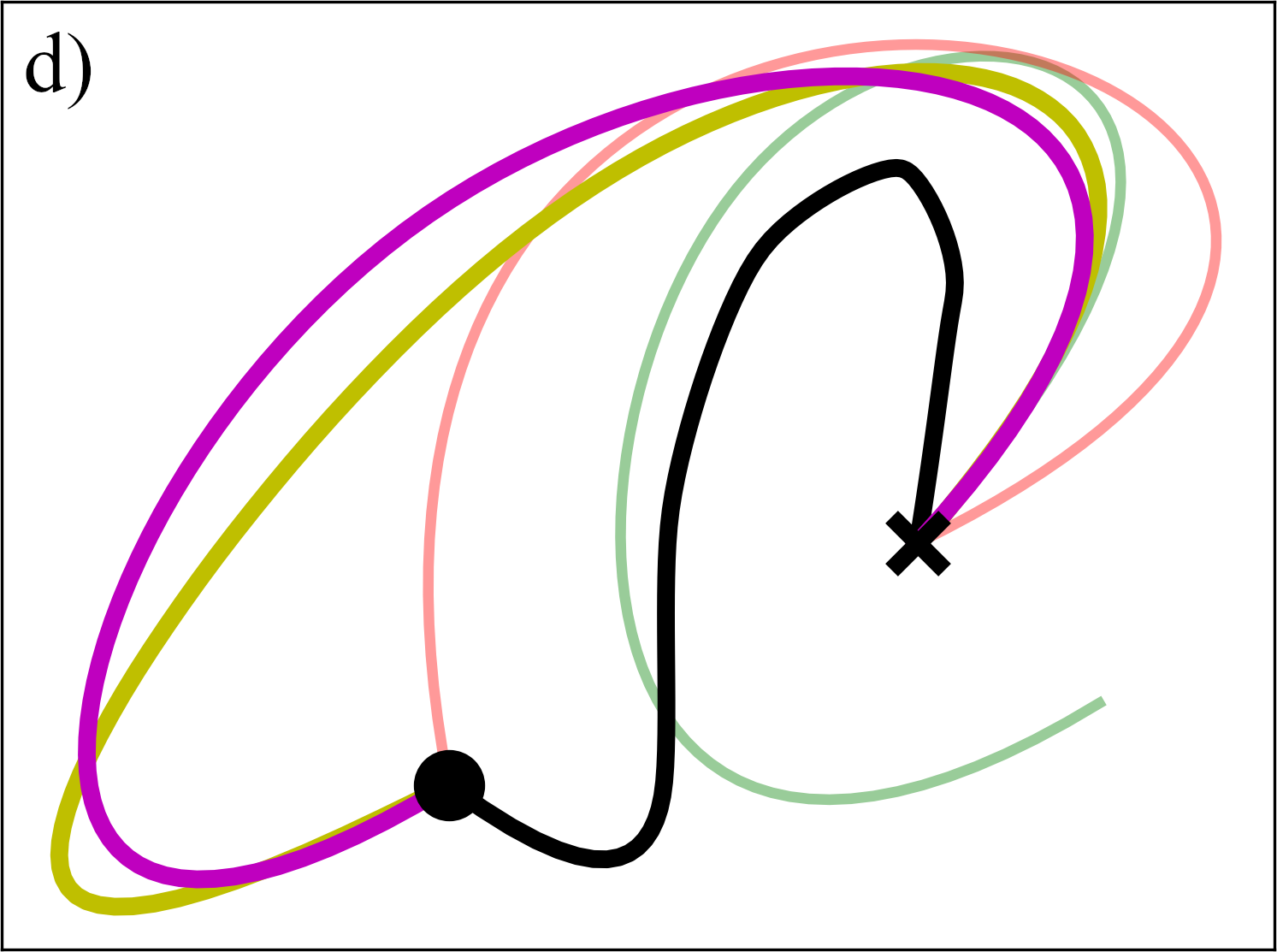}
\includegraphics[trim=0 0em 0 0, clip, width=0.98\textwidth]{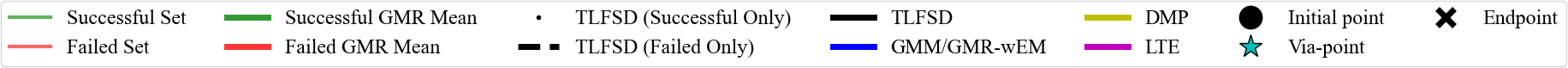}
\caption{\small{(a) results of using various combinations of failed and successful demonstrations in the presence of constraints. (b) and (c) comparison between TLFSD and GMM/GMR-wEM in a simulated pushing task. (d) comparison between DMPs, LTE, and TLFSD in a simulated writing task.}} \label{2d_exps}
\end{figure*}

We evaluated TLFSD through multiple 2D and 3D real-world experiments. For 2D experiments, we captured demonstrations through a GUI using a computer mouse, although other pointing devices could also be used. For 3D experiments, we captured demonstrations on a 6-DOF UR5e manipulator arm through kinesthetic teaching. All the captured raw demonstrations in each set were smoothed, temporally aligned, and labeled as successful or failed. The reproduced trajectories in each experiment were executed on the manipulator to verify the feasibility of the proposed approach. Note that all figures in this section use the legend presented in Fig.~\ref{2d_exps}.

\subsection{Modeling using Failed and Successful Demonstrations}
\label{sec:Exp_A}

In the first experiment, we show that our approach can deal with 
\begin{enumerate*}[label=(\roman*)]
  \item an empty failed set, 
  \item an empty successful set, and 
  \item when both sets include demonstrations of the skill.
\end{enumerate*}
As illustrated in Fig.~\ref{2d_exps}(a), we recorded and labeled two successful and two failed demonstrations. The statistical mean for each set was also estimated as explained in Sec.~\ref{sec:encoding} and plotted. It can be seen that all the demonstrations approach the same final point. However, to show the generalization capabilities of our approach, we conditioned our algorithm to an initial point which is close to the failed mean, a via-point, and the common endpoint. 

The first reproduction (dotted) was generated using only successful demonstrations, assuming the failed set was empty. This reproduction satisfies three spatial constraints while following the GMR mean of successful demonstrations. Since this trajectory does not use the information from the failed demonstration set, it passes through the failed set from the initial point to the via-point and hence could be considered incorrect by the user. The next reproduction (dashed) was generated using only the failed demonstration set, assuming the successful set was empty. This reproduction avoids the failed demonstrations while satisfying the constraints but cannot imitate the desired shape of the skill without knowledge of the successful demonstration set. The final reproduction (solid) uses both failed and successful demonstration sets. Since the initial point is close to the failed set, the algorithm reproduces an optimum solution that diverges from the failed set while at the same time satisfying the shape of the successful set and the given constraints. 

\subsection{Importance of Failed Demonstrations}
\label{sec:Exp_B}
This experiment shows the importance of considering failed demonstrations in a specific task where a reaching skill was demonstrated in the presence of an obstacle. We assume that the obstacle is unknown and undetectable by the robot. It can be seen in Fig.~\ref{fig1} that the user has given two demonstrations that avoid the obstacle (successful set) and two demonstrations that collide with the obstacle (failed set). Our results show that given only the successful demonstration set, the algorithm generates a trajectory that tries to preserve a mean between the two demonstrations and hence collides with the obstacle. However, when the algorithm uses information from both the failed and successful demonstrations, it learns to generate trajectories that avoid the obstacle. It should be noted that we do not rely on any object detection or obstacle avoidance method, instead assuming that the information about the obstacle was only provided through failed demonstrations. Thus, to guarantee that the reproductions of the skill avoid the obstacle, the given failed set should cover the boundaries of the obstacle. Alternatively, different solutions can be found by tuning the hyper-parameters of the algorithm accordingly.

\subsection{Comparing to GMM/GMR-wEM}

In this experiment, we compared TLFSD against Gaussian Mixture Models/Gaussian Mixture Regression with weighted Expectation Maximization (GMM/GMR-wEM)~\cite{argall2010tactile}, which was designed for iterative skill refinement. GMM/GMR-wEM enables the user to refine a skill, by assigning higher weights to newer (more optimal) demonstrations and forgetting older (sub-optimal) ones. We conducted an experiment on a pushing skill for closing a drawer as can be seen in Fig.~\ref{2d_exps}(b) and \ref{2d_exps}(c). This task has an implicit constraint that requires the trajectory to remain horizontal when pushing the drawer. Note that we did not impose this constraint on the optimization problem and expect the algorithm to learn it from demonstrations.
As depicted in Fig.~\ref{2d_exps}(b), we recorded three demonstrations that fail to complete the task because of an upward movement at the end of the trajectory and one successful demonstration. Given these demonstrations sets, TLFSD is able to find a reproduction that successfully completes the task without any iteration. 

We then repeated the experiment using GMM/GMR-wEM by assigning a higher weight to the successful demonstration and smaller weights to failed ones (GMM/GMR-wEM weights cannot be negative). It can be seen that GMM/GMR-wEM was not able to pull the regression mean towards the successful demonstration. After adding two additional successful demonstrations, as shown in Fig.~\ref{2d_exps}(c), the reproduced trajectory using GMM/GMR-wEM was able to successfully close the drawer.

\begin{figure*}[h!]
\centering
\includegraphics[trim=0 0em 0 0, clip, width=0.22\textwidth, height=0.34\columnwidth]{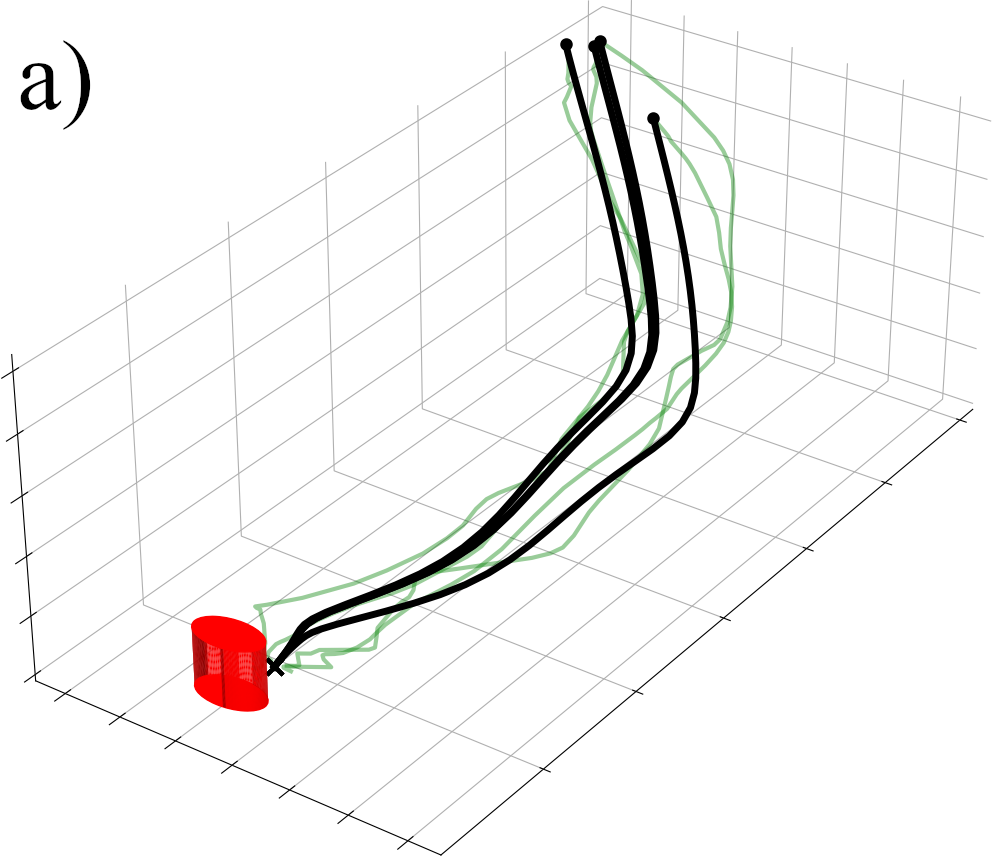}
\includegraphics[trim=0 0em 0 0, clip, width=0.22\textwidth, height=0.34\columnwidth]{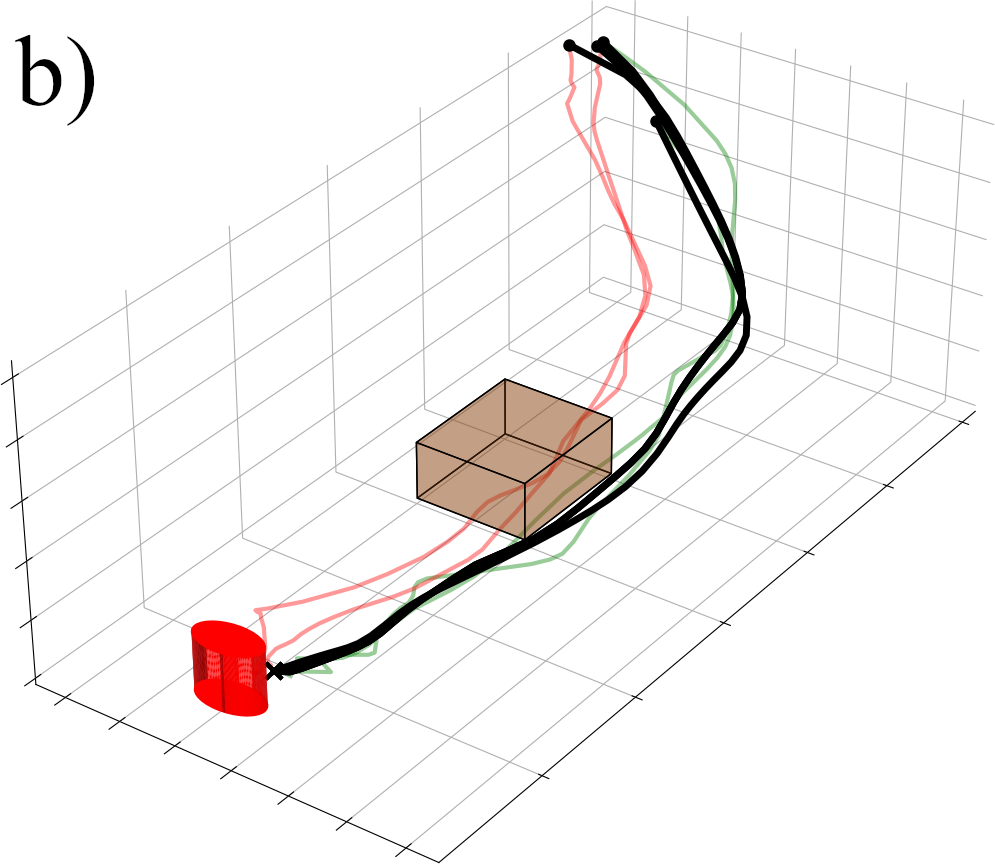}
\includegraphics[trim=0 0em 0 0, clip, width=0.005\columnwidth, height=0.34\columnwidth]{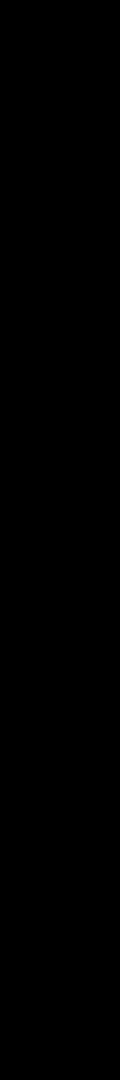}
\includegraphics[trim=0 0em 0 0, clip, width=0.22\textwidth, height=0.34\columnwidth]{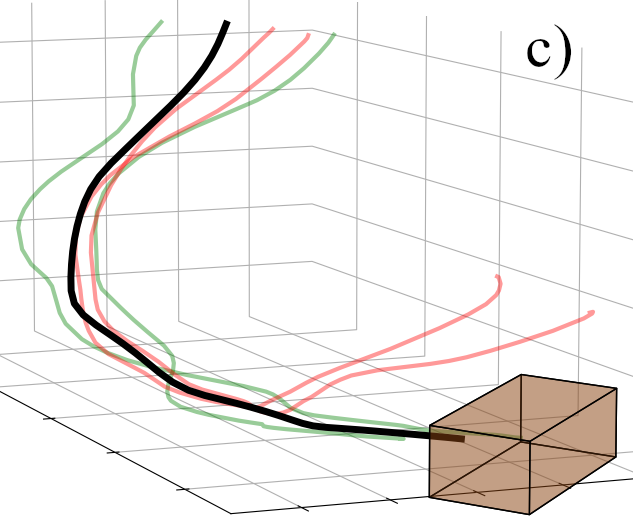}
\includegraphics[trim=0 0em 0 0, clip, width=0.22\textwidth, height=0.34\columnwidth]{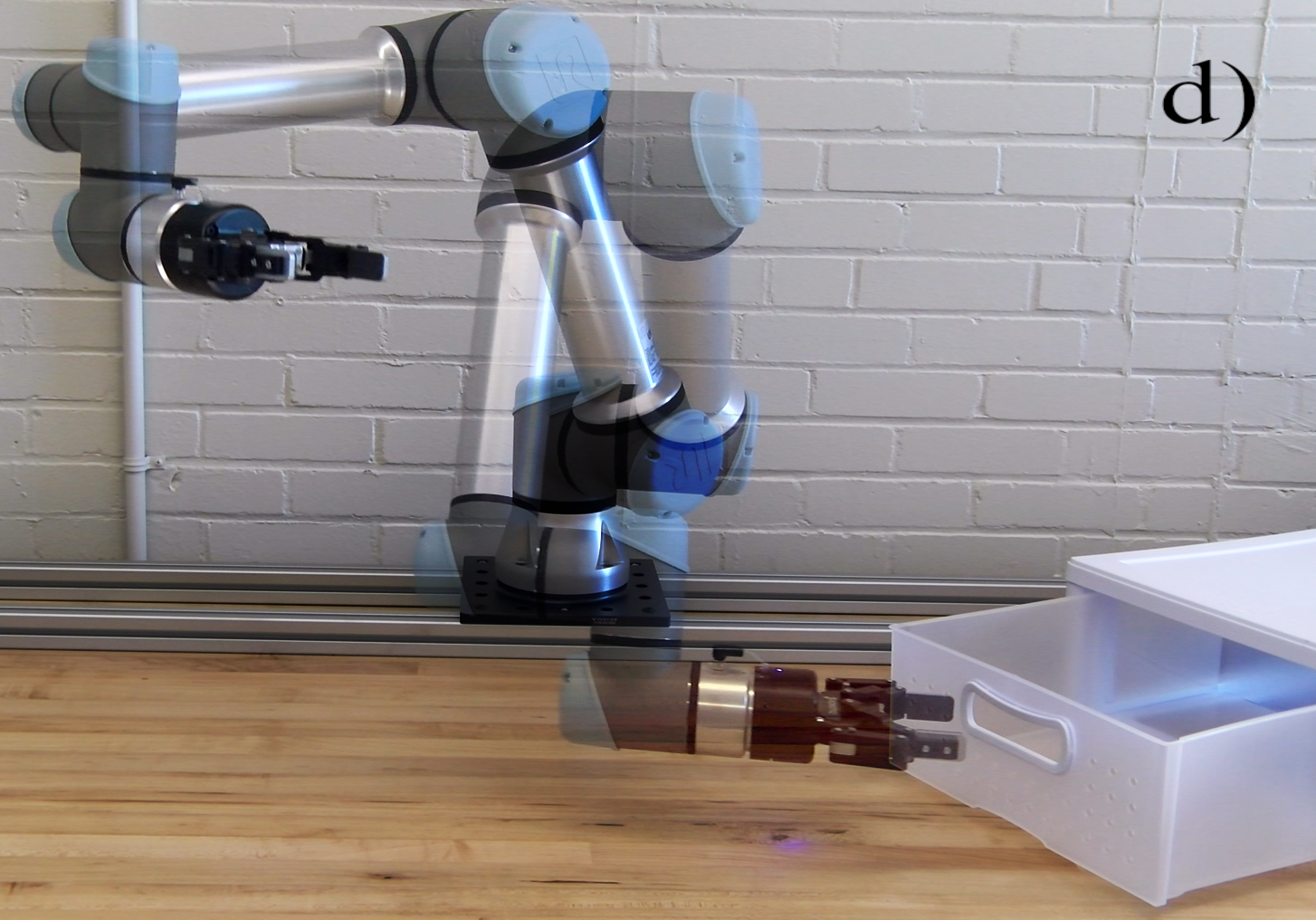}
\caption{\small{(a) and (b) TLFSD adapting to the change in scene when an obstacle is placed, resulting in successful reproductions for a reaching task. (c) and (d) TLFSD finds a successful reproduction for a pushing task with no constraints given.}} \label{3d_demo}
\end{figure*}

\begin{table}[h]
\centering
\caption{\small{Comparing TLFSD to DMP and LTE across various similarity metrics (bold numbers represent best performances).}}
\label{tab:1D_compare}
\begin{tabular}{lccc} \toprule
 & SSE & SEA & CRV \\ \midrule 
TLFSD & \textbf{29.2} & \textbf{2.1} &  15.5   \\ 
DMP   &        140.4  &         4.9  &  9.1    \\ 
LTE   &        154.0  &         5.6  & \textbf{1.0}  \\ \bottomrule
\end{tabular}
\end{table}

\subsection{Comparing to Conventional LfD representations}
Additionally, we compared TLFSD against two conventional single-demonstration LfD representations: Dynamic Movement Primitives (DMPs)~\cite{pastorDMP2009} and Laplacian Trajectory Editing (LTE)~\cite{Nierhoff2016LTE}. Neither DMPs nor LTE has been designed to encode failed demonstrations, and they are only given information of successful demonstrations. As depicted in Fig.~\ref{2d_exps}(d), a single successful and a single failed demonstration of a 2D writing skill were provided. The endpoints of the two demonstrations are the same, while the initial point during reproduction is constrained to the initial point of the failed demonstration. We use three dissimilarity metrics: Sum of Squared Errors (SSE), Swept Error Area (SEA)~\cite{Khansari-Zadeh2014SEA}, and SSE in curvature space (CRV). These metrics were selected as they each measure different properties: SSE measures spatial and temporal displacement, SEA measures spatial displacement invariant of time, and CRV measures the difference in curvature between the two trajectories. Fig.~\ref{2d_exps}(d) depicts reproductions using DMP, LTE, and TLFSD. Results of the comparison between these reproductions and the successful demonstration are reported in Table~\ref{tab:1D_compare}. Based on SSE and SEA, TLFSD generates a reproduction with higher similarity in Cartesian space. In curvature space, however, TLFSD is the most dissimilar of the three reproductions which may be considered as failure by the user, especially in a writing task where shape curvature must be preserved. Based on CRV, the LTE reproduction is best, because LTE encodes the skill in Laplacian coordinates and preserves shape properties including curvature. In Section~\ref{coords_extension}, we propose an extension of TLFSD to address this shortcoming.

\begin{figure}[b]
\centering
\includegraphics[trim=0 0em 0 0, clip, width=0.98\columnwidth]{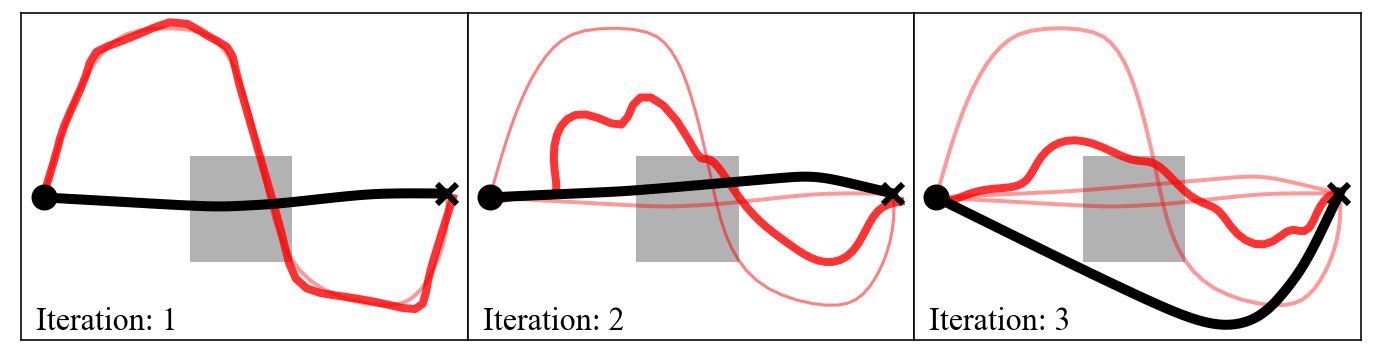}
\caption{\small{Iterative process of finding a solution that avoids the (unknown) obstacle and reaches the end point only from failed demonstrations. A solution was found after three iterations.}} \label{iter_demo}
\end{figure}

\subsection{Iterating From Failure to Success using Human Input}

In this experiment, we employ TLFSD in an iterative learning manner to find a successful reproduction when only failed demonstrations are given. As it can be seen in Fig.~\ref{iter_demo}(left) a single demonstration of a skill was provided which collides with an obstacle. The algorithm was not given any information regarding the obstacle, only the demonstration was labeled as failed. The first reproduction in Fig.~\ref{iter_demo}(left) that satisfies the initial and final conditions also collided with the obstacle and was labeled as failed by the user. The failed reproduction then was added as a new demonstration in the failed set. As shown in Fig.~\ref{iter_demo}(middle) the algorithm incorporates the new observation into the model and generates a `better' reproduction. After three iterations, the algorithm generates a reproduction which does not collide with the obstacle, resulting in a successful reproduction of the skill as seen in Fig.~\ref{iter_demo}(right). This experiment shows that TLFSD is able to reproduce better trajectories (i.e., trajectories that diverge from failed demonstrations) iteratively using labels provided by the human user.

\subsection{Real-World Experiments}
\label{sec:exp_real}
We tested TLFSD in a real-world 3D reaching task in an environment similar to Fig.~\ref{fig1}(right). As illustrated in Fig.~\ref{3d_demo}(a), first we captured four demonstrations and labeled them as successful. Four reproductions of the skill using TLFSD from different initial points are plotted. All reproductions were able to successfully achieve the goal of the skill. Then, the box shown in Fig.~\ref{3d_demo}(b) was added to the scene as an obstacle that intersected with some of the given demonstrations which were then labeled as failed. New TLFSD reproductions from the initial point of each given demonstration were generated, all of which adapted to be able to complete the task and avoid the obstacle\footnote{Accompanying video at: \url{https://youtu.be/sRdOm_9nJ8g}}.

Additionally, we tested a pushing skill in 3D similar to the skill shown in Fig.~\ref{2d_exps}(b) and \ref{2d_exps}(c). In this experiment, as shown in Fig.~\ref{3d_demo}(c) and \ref{3d_demo}(d), two successful and two failed demonstrations were given. The two failed demonstrations were unable to complete the skill successfully as they raise in elevation before fully closing the drawer. The TLFSD reproduction is able to complete the task by encoding the important implicit features of the movement from  successful and failed demonstrations. Fig.~\ref{3d_demo}(d) shows the execution of the successful reproduction on a UR5e manipulator.

\section{Extension into other Coordinates} 
\label{coords_extension}

Sometimes the reason for a successful or failed demonstration may not be captured in Cartesian coordinates alone. As shown by Ravichandar et al., some features that represent shape properties (e.g., curvature), might be represented better in other differential coordinates such as Tangent or Laplacian~\cite{Ravichandar2019MCCB}. Their approach encodes information from demonstrations in multiple coordinates and therefore can model skills by finding and replicating different types of trajectory features. Inspired by this idea, we extend our approach by including Tangent and Laplacian coordinates in addition to Cartesian coordinates. We transform demonstrations captured in Cartesian coordinates $^\mathcal{C}\vect{X}$ into the Tangent and Laplacian coordinates using $^\mathcal{G}\vect{X} = \vect{G}^\mathcal{C}\vect{X}$ and $^\mathcal{L}\vect{X} = \vect{L}^\mathcal{C}\vect{X}$ operations where $\vect{G}$ and $\vect{L}$ represent the Tangent and Laplacian coordinate matrix, respectively defined as
\begin{equation}
\vect{G} =
\left[\begin{smallmatrix}
 -1 & 1 &  0 &  \cdots & 0 \\
0 &  -1 & 1 & \cdots & 0 \\
\vdots & \ddots & \ddots & \ddots & \vdots \\
0 & \cdots & 0 &  -1 & 1\\
0 & \cdots & \cdots & 0 & -1   \\
\end{smallmatrix}\right],
\vect{L} = \frac{1}{2}
\left[\begin{smallmatrix}
 2 & -2 &  0 & \cdots & \cdots & 0 \\
-1 &  2 & -1 & 0 & \cdots & 0 \\
0 & -1 &  2 & -1 & \cdots  & 0 \\
\vdots & \ddots & \ddots & \ddots & \ddots & \vdots \\
0 & \cdots & 0 & -1 &  2 & -1\\
0 & \cdots &  \cdots & 0 & -2 & 2   \\
\end{smallmatrix}\right]. \nonumber
\end{equation}

Given the set of demonstrations in three coordinates $ \{ ^\mathcal{C}\vect{X},  ^\mathcal{G}\vect{X},  ^\mathcal{L}\vect{X} \} $, we construct an new optimization problem similar to \eqref{eq:opt2} by extending the quadratic cost $J^Q$ to all coordinates and combine them as follows
\begin{align}
    &\underset{\vect{X} \in \mathbb{R}}{\text{minimize }} \sum_k \alpha_k J^Q(^k\vect{X}) + J^R(^{\mathcal{C}}\vect{X}) + \frac{1}{2 \beta} \sum_{i \in \mathcal{E}} C_i^2(^{\mathcal{C}}\vect{X}), \label{eq:opt3}
     \nonumber
\end{align}

\noindent where $^k\vect{X}$ denotes a transformed demonstration with $k \in \{\mathcal{C, G, L}\}$ representing the target coordinate, and  $\alpha_\mathcal{C}$, $\alpha_\mathcal{G}$ and $\alpha_\mathcal{L}$ are scaling factors that represent importance for the Cartesian, Tangent, and Laplacian coordinates, respectively. The $\alpha_k$ values can be auto-tuned using meta-optimization similar to \cite{Ravichandar2019MCCB}. The extension to multiple coordinates allows our approach to encode different types of skill features within both failed and successful demonstration sets.

\begin{figure}[b]
\centering
\includegraphics[trim=0 0em 0 0, clip, width=0.55\columnwidth]{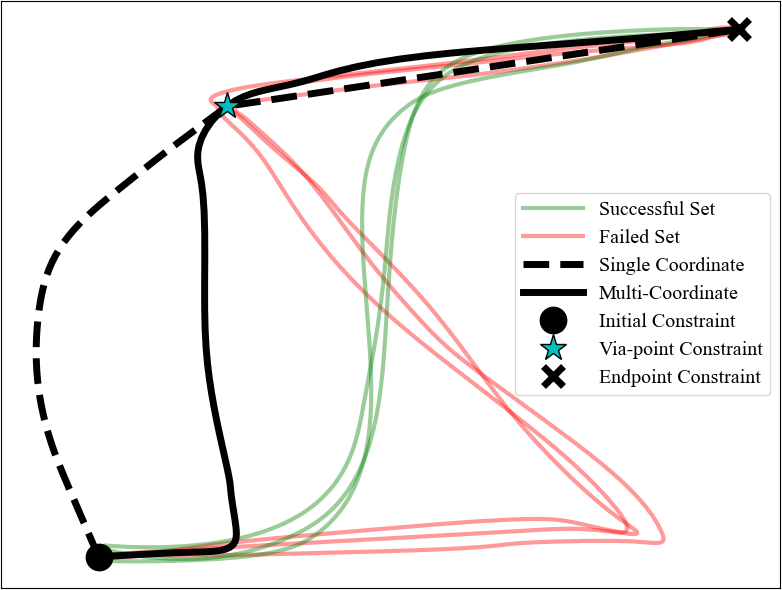}
\caption{\small{Encoding in multiple coordinates (solid black) compared to encoding only in Cartesian coordinate (dashed black). }} \label{coords_demo}
\end{figure}

We evaluate the extended approach and show its effectiveness compared to the original algorithm by conducting an experiment with two sets of demonstrations as shown in Fig.~\ref{coords_demo}. Compared to the reproduction only based on Cartesian coordinate encoding (dashed black), the reproduction based on multi-coordinate encoding (solid black) preserves curvature and tangent features of the successful demonstration set while avoiding failed demonstrations. Although both reproductions satisfy all the task constraints (initial, final, and via-points), the multi-coordinate reproduction also satisfied the implicit constraints to generate a more similar trajectory. 

\section{Future Work}
\label{sec:Conclusion}

We have proposed a novel LfD approach that learns from both successful and failed demonstrations of a skill. Possible future work includes incorporating techniques such as skill refinement into the learning process, as well as improving the efficiency of the learning process. Alternatively, methods for improving the learning process could consider semantics, with the intent of understanding why a demonstration failed to better model failure and avoid reproducing it.

\typeout{}
\bibliographystyle{IEEEtran}
\bibliography{references}

\end{document}